\documentclass[lettersize,journal]{IEEEtran}
\usepackage{amsmath,amsfonts}
\usepackage{algpseudocode}
\usepackage{array}
\usepackage{subcaption}
\usepackage{textcomp}
\usepackage{stfloats}
\usepackage{url}
\usepackage{verbatim}
\usepackage{algorithm}
\usepackage{amssymb}
\usepackage{graphicx}
\usepackage{cite}
\usepackage{multirow}
\usepackage{color}
\usepackage{booktabs}
\usepackage{hyperref}
\hypersetup{
    colorlinks=true,
    linkcolor=blue,
    filecolor=magenta,      
    urlcolor=cyan,
    pdftitle={NanoNAS on IoT},
    pdfpagemode=FullScreen,
    }
\usepackage{tikz}

\begin{document}

%\title{Automatic Design of Neural Networks at the Edge}
\title{Searching Neural Architectures for Sensor Nodes on IoT Gateways}

\author{Andrea~Mattia~Garavagno,~\IEEEmembership{Graduate Student Member,~IEEE,} 
Edoardo~Ragusa,~\IEEEmembership{Member,~IEEE,} Antonio~Frisoli,~\IEEEmembership{Member,~IEEE,} Paolo~Gastaldo

        % <-this % stops a space
\thanks{Manuscript received Month Day, Year; revised Month Day, Year; (Corresponding author: Andrea Mattia Garavagno.)}
\thanks{Andrea Mattia Garavagno, Edoardo Ragusa, and Paolo Gastaldo are with the Department of Electrical, Electronic, Telecommunications Engineering and Naval Architecture, University of Genoa, 16126 Genoa, Italy (e-mail: AndreaMattia.Garavagno@edu.unige.it, Edoardo.Ragusa@unige.it, Paolo.Gastaldo@unige.it).}
\thanks{Andrea Mattia Garavagno and Antonio Frisoli are with the Institute of Mechanical Intelligence, Scuola Superiore Sant’Anna, Pisa 56127, Italy (e-mail: AndreaMattia.Garavagno@santannapisa.it, Antonio.Frisoli@santannapisa.it).}}

\markboth{IEEE Internet of Things Journal,~Vol.~xx, No.~x, Month~Year}%
{Shell \MakeLowercase{\textit{et al.}}: A Sample Article Using IEEEtran.cls for IEEE Journals}

%
%\IEEEpubid{This work has been submitted to the IEEE for possible publication. Copyright may be transferred without notice, after which this version may no longer be accessible.}

\maketitle

\begin{tikzpicture}[remember picture,overlay]
\node[anchor=south,yshift=15pt] at (current page.south) {
    \parbox{\textwidth}{
        \centering \scriptsize
        This work has been accepted for publication in IEEE Internet of Things Journal. Final publication is available at \url{https://doi.org/10.1109/JIOT.2025.3581442} .
    }
};
\end{tikzpicture}

\begin{abstract}
This paper presents an automatic method for the design of Neural Networks (NNs) at the edge, enabling Machine Learning (ML) access even in privacy-sensitive Internet of Things (IoT) applications. The proposed method runs on IoT gateways and designs NNs for connected sensor nodes without sharing the collected data outside the local network, keeping the data in the site of collection. This approach has the potential to enable ML for Healthcare Internet of Things (HIoT) and Industrial Internet of Things (IIoT), designing hardware-friendly and custom NNs at the edge for personalized healthcare and advanced industrial services such as quality control, predictive maintenance, or fault diagnosis. By preventing data from being disclosed to cloud services, this method safeguards sensitive information, including industrial secrets and personal data. The outcomes of a thorough experimental session confirm that --on the Visual Wake Words dataset-- the proposed approach can achieve state-of-the-art results by exploiting a search procedure that runs in less than 10 hours on the Raspberry Pi Zero 2. 
\end{abstract}

\begin{IEEEkeywords}
Neural Architecture Search, Edge AI, Healthcare Internet of Things, Industrial Internet of Things.
\end{IEEEkeywords}

\section{Introduction}
Neural Networks (NNs) are widely used in Internet of Things (IoT) applications \cite{dong2023graph}. In this context, often the data collected by the available sensors are added to the training set with the purpose of improving generalization performances. On the other hand, in some cases, the data can be sensitive; healthcare data \cite{bhuiyan2021internet}, industrial data \cite{chen2024knowledge} and biometric data \cite{wu2020ecg} provide possible examples. Privacy concerns prevent some entities from accessing the benefits of machine learning (ML), as they may be unable or unwilling to share their data with cloud services that can train or even automatically design a custom neural network (NN) \cite{li2018privacy}.

To overcome this problem and make NNs available even in privacy-sensitive applications, we provide a method to automatically design NNs for sensor nodes on IoT gateways using locally collected data. This in turn means that the whole process can be completed without any access to cloud services, therefore maintaining data privacy. 
In the case of Healthcare Internet of Things (HIoT), this approach can be used to provide personalized healthcare services with customized and hardware-friendly NNs for each patient \cite{habibzadeh2019survey, aceto2020industry}.
Industrial Internet of Things (IIoT) can also benefit from this approach. For example, one can provide hardware-friendly and customized NNs for intelligent fault diagnosis (IFD) of production machinery or quality control, without facing the risk of data leakage that could reveal industrial secrets \cite{serror2020challenges}.

\IEEEpubidadjcol

\begin{figure}
    \centering
    \includegraphics[width=0.5\textwidth]{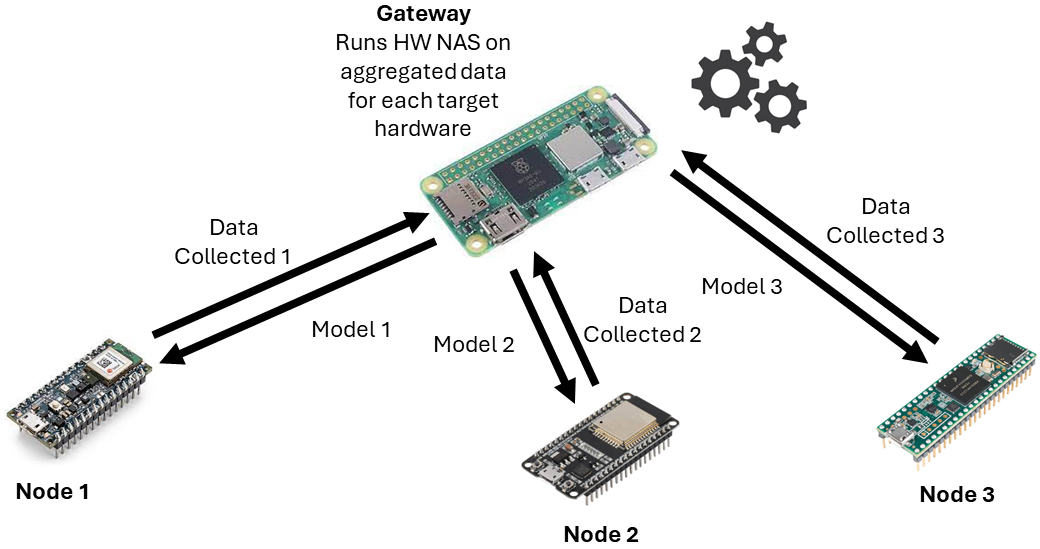}
    \caption{Automatic designing of neural architectures at the edge, running HW-NAS on an IoT gateway, using the locally collected data.}
    \label{fig:proposed_approach}
\end{figure}

A synopsis of the proposed scenario is provided in Fig. \ref{fig:proposed_approach}. First, the data collected by the sensor nodes are aggregated by the local gateway; then, the gateway adopts the proposed method to design a custom NN for each connected sensor node, as, in general, each node may differ from the other in terms of hardware resources. Finally, the designed NNs are deployed on the sensor nodes. The whole process can be completed without transferring data to any cloud infrastructure, thus preserving data privacy. 

The proposed method exploits Hardware-Aware Neural Architecture Search (HW-NAS) to automatically design NNs for sensor nodes; such an approach is becoming popular in IoT applications \cite{ren2021comprehensive, zhang2022toward, wang2020neural}. Still, this work addresses a critical issue: the computational cost of the search procedure. The general goal of state-of-the-art HW-NAS is to generate NNs that fit the resources available on the hardware that will run the inference process. However, the resources available on the platform that runs the search procedure itself are usually considered unconstrained \cite{hwnas_survey}. The latter assumption becomes critical when working with constrained devices such as IoT gateways \cite{adhinugraha2020internet, lu2021adaptive}. If one wants to avoid any data transfer to computing resources that are external to the IoT network, then the search procedure should be designed to run on the available IoT gateway.

The key challenges that arise when running HW-NAS on edge platforms are a) dealing with the low computational capabilities and b) dealing with the low memory availability of these devices. To address these challenges, we adopt a derivative-free search strategy combined with a regular search space. The regular search space eases the evaluation process by employing architectures that are hardware-friendly in the sense that they do not feature too many branches to be followed during the backpropagation phase, saving memory and computations. The derivative-free search strategy avoids the usage of derivatives and therefore requires less memory and less computations compared to derivative-based algorithms. 

The present paper inherits and expands the work recently presented in \cite{garavagno2024running}, further developing the idea of running HW-NAS on constrained devices. The main contributions of this paper can be summarized as follows.

\begin{itemize}
    \item A novel search strategy that obtains state-of-the-art results on the Visual Wake Words dataset while reducing search time from 4 days to 10 hours when running on the Raspberry Pi Zero 2.
    \item An adaptation mechanism which tunes the search space according to the energy and time available to the IoT gateway, enabling the execution of HW-NAS even in the case of a limited energy budget or with a limited amount of execution time at disposal.
    \item The added capability of working with time series, obtaining state-of-the-art results on the CWRU dataset, a dataset for Intelligent Fault Diagnosis, in just 2 hours and 52 minutes when running on a Raspberry Pi 4. 
    \item The release of the proposed HW-NAS as open-source software at \url{https://github.com/AndreaMattiaGaravagno/GatewayNAS}; the software is designed for embedded devices based on Linux. 
\end{itemize}

The experimental sessions involved three different single-board computers typically used for implementing IoT gateways and three different ultra-low-power microcontroller units (MCUs) as reliable instances of sensing nodes. The experiments showed that the proposed HW-NAS could successfully meet the constraints on execution time and energy consumption while generating architectures that scored state-of-the-art performances on the selected benchmarks.

The rest of the paper is organized as follows. Section \ref{sec:related} reviews related works. Section \ref{sec:background} briefly introduces the reader to the basics of HW-NAS and then focuses on the specific problem of running HW-NAS on resource-constrained devices. Section \ref{sec:proposal} presents the proposed approach to the design of an HW-NAS that fits the requirements of the IoT environment. Section \ref{sec:exp_setup} reports on the experimental setup, while Section \ref{sec:experiments} presents the outcomes of the experiments. Finally, concluding remarks are provided in Section \ref{sec:conclusion}.

\section{Related Works} \label{sec:related}
HW-NAS represents a valuable solution for automatically designing deep learning models whose inference phase should run on IoT sensor nodes \cite{zhang2020empowering, guo2024semantic, lin2024multi}. Wang et al. \cite{wang2020neural} applied HW-NAS to the design of NN that must be robust against adversarial attacks in 6G-enabled AI-enabled Internet-of-Things (AIoT) systems. In \cite{prabakaran2021bionetexplorer} an HW-NAS generates and explores multiple NN architectures for biosignal processing in wearable devices. Du et al. \cite{du2022ssvep} exploited HW-NAS to design tiny deep learning models to analyze the emotional information of steady-state visual evoked potential signals in the IoT framework. Dissem et al. \cite{dissem2024neural} used HW-NAS to design optimal autoencoders to perform anomaly detection in smart buildings. In \cite{ragusaIOT}, HW-NAS supports the development of IoT systems for vibration damage detection.

\begin{table}[!h]
    \centering
    \caption{Target execution platform of state-of-the-art HW-NAS.}
    \label{tab:sota}
    \begin{tabular}{c | c c c c}
        \toprule
        \multirow{2}{*}{Work}                     & \multirow{2}{*}{GPU} & \multirow{2}{*}{CPU} & \multicolumn{2}{c}{Embedded}  \\
                                                  &                      &                      & Static  & Adaptive            \\ \hline
        Mnasnet \cite{tan2019mnasnet}             & x                    &                      &         &                     \\
        MCUNet \cite{lin2020mcunet}               & x                    &                      &         &                     \\
        Micronets \cite{banbury2021micronets}     & x                    &                      &         &                     \\
        ColabNAS \cite{garavagno2024colabnas}     & x                    &                      &         &                     \\
        NanoNAS v1 \cite{garavagno2023hardware}   &                      & x                    &         &                     \\
        NanoNAS v2 \cite{garavagno2024affordable} &                      &                      & x       &                     \\
        Proposal                                  &                      &                      &         & x                   \\ \bottomrule
    \end{tabular}
\end{table}

Originally, HW-NAS was a resource-hungry procedure; as a result, one needed expensive computing facilities to run it. In the last years, though, efforts have been made to decrease the computational cost of HW-NAS procedures. From the 40,000 GPU hours required by Mnasnet \cite{tan2019mnasnet}, the execution time was reduced to 300 GPU hours by MCUNet \cite{lin2020mcunet}. In addition, custom techniques leveraging derivative-free optimization and hand-crafted search spaces have been introduced to meet the resource constraints of ultra-low-power hardware: in \cite{garavagno2024colabnas} the search cost was 4 GPU hours, while Garavagno et al. eventually removed the need for a GPU \cite{garavagno2023hardware}. Indeed, two recent works showed that it can be feasible to run the HW-NAS itself on an embedded system \cite{garavagno2024running, garavagno2024affordable}. The main idea is to add a constraint on the memory available on the platform running the search procedure. Table \ref{tab:sota} offers an overview of the state-of-the-art HW-NAS; the works are classified according to the computing resources needed to perform the search procedure.

The literature provides other approaches that can be adopted to adapt the architecture of a DNN to changes in the deployment environment.  
Dynamic Neural Networks enjoy favorable properties that are absent in common neural networks \cite{skarding2021foundations}: they can adapt their topology and parameters in response to varying input during inference; 
they can learn a set of diverse features, enhancing their generalization capabilities; they support interpretable ML, giving the ability to understand which part of the network has generated the output. However, the known gap between theoretical and practical efficiency makes them unsuitable for modern computing devices \cite{han2021dynamic}. %sezione 7.4 della citazione

Domain adaptation \cite{pan2009survey} transfers knowledge from a source domain to a target domain having a different data distribution. It has been successfully applied to increase the generalization capability of NNs for IoT applications trained in a situation of data scarcity \cite{yang2019learning, zou2018robust, yu2022intelligent, santamato2024leveraging}. Nonetheless, the burden of designing a neural architecture compatible with the hardware constraint of the sensor node remains on the user.

Hence, the present paper wants to go one step further in the design of HW-NAS that can run at the edge of IoT systems, without relying on external resources. The proposed adaptive HW-NAS can automatically adjust the search procedure according to the constraints on execution time and energy budget. This in turn means that it can run on generic gateways in an IoT network. 

\section{Background} \label{sec:background}
\subsection{Hardware-Aware Neural Architecture Search}
HW-NAS automatizes the process of finding the best neural architecture for solving a given task on a target hardware. HW-NAS is typically cast as an \textbf{optimization problem} constrained by the resources available on the target hardware. Its objective function defines the set of metrics adopted for selecting the best neural architecture among candidate solutions, i.e., the \textbf{evaluation process} \cite{benmeziane2021comprehensive}. Equation (\ref{eq:problem_old_old}) shows an example of the typical optimization problem, where one wants to find the architecture $A$ having the maximum validation accuracy, $max \Gamma(A)$, among those meeting a set of constraints. In this example the constraints are the amount of RAM ($\xi_{RAM}$), the amount of Flash memory ($\xi_{Flash}$), and the multiply and accumulate (MAC) instructions ($\xi_{MAC}$); accordingly, $\phi_{RAM}(A)$, $\phi_{Flash}(A)$, and $\phi_{MAC}(A)$ refer, respectively, to the RAM usage, the Flash memory usage, and the MAC instructions of the candidate architecture $A$ on the deployment device. 

\begin{equation} \label{eq:problem_old_old}
\begin{aligned}
\begin{cases}
   \hfil \max \Gamma(A)\\
   \hfil \phi_{RAM}(A) \leq \xi_{RAM} , \phi_{Flash}(A) \leq \xi_{Flash} \\
   \hfil \phi_{MAC}(A) \leq \xi_{MAC} \\
   \hfil \xi_{RAM}, \xi_{Flash}, \xi_{MAC}  > 0
\end{cases}
\end{aligned}
\end{equation}

The set of candidate solutions, also known as \textbf{search space}, is defined by fixed architectural rules determining which NNs are allowed. Rules can be defined with different granularity: they can act on layers (layer-wise search space), on a group of layers, often referred as cells (cell-based search space), or both (hierarchical search space) \cite{benmeziane2021comprehensive}. 

The optimization technique adopted for solving the problem, i.e., the \textbf{search strategy}, defines how the search space is explored when looking for the best neural architecture \cite{benmeziane2021comprehensive}. In the first place, techniques based on Reinforcement Learning (RL) \cite{tan2019mnasnet}, Evolutionary Aging (EA) \cite{cai2019once}, Random Search \cite{li2020random}, and Bayesian Optimization (BO) \cite{zoph2018learning} where adopted given the non-smoothness of the HW-NAS problem. Then, a novel formulation of HW-NAS \cite{liu2018darts} allowed the usage of gradient-based techniques \cite{cai2018proxylessnas, wu2019fbnet} for solving the optimization problem. Nonetheless, the majority of search strategies are known to be resource intensive \cite{benmeziane2021comprehensive} and can require up to tens of thousands hours of execution on specialized hardware \cite{lin2020mcunet}. In the recent years, though, new approaches have been introduced that lower the search cost of HW-NAS by adopting custom search strategies \cite{zhang2020fast, garavagno2023hardware, garavagno2024running}.

\subsection{Running HW-NAS on Resource Constrained Devices}\label{sec:search_space}
Running HW-NAS on embedded devices poses major challenges, as limited resources are available. Two recent works \cite{garavagno2024running} \cite{garavagno2024affordable} have shown that the search cost can be significantly reduced by exploiting a custom derivative-free optimization technique and a cell-based search space; in both cases the HW-NAS was designed to generate NNs that should run on ultra-low-power computing platform. 

The adopted search space \cite{garavagno2024running} builds tiny Convolutional Neural Networks (CNNs) stacking four different types of cells: a pre-processing cell, a base cell, a building cell, and a classifier cell. The pre-processing cell applies min-max standardization to improve gradient descent convergence rate \cite{shanker1996effect}. The base cell is a single convolutional layer with $k$ kernels acting as the foundation on which building cells will be stacked. A single building cell first halves the resolution of the extracted features using the max pooling operator, then applies convolution followed by batch normalization to stabilize further and speed up the training \cite{ioffe2015batch}; then a rectified linear unit activation builds the new feature maps. The number of kernels $n_c$ of a convolutional layer depends on $k$ and on the number of cells previously stacked $c$, and can be computed as $n_{c} = n_{c - 1} + 2^{1-c} n_{c - 1}$, where $n_{0} = k$ \cite{garavagno2024running}. The latter formula is inspired by VGG16 \cite{simonyan2014very}, where instead of doubling the kernels after each block, we gradually decrease the increment to save resources.

Finally, the classifier cell reduces the extracted features by applying global average pooling to improve the model’s generalization capability \cite{lin2013network}; a fully connected layer produces the class predictions, using softmax activation and a number of neurons equal to the number of classes. All the kernels use a $3$ by $3$ receptive field.

Therefore, in the adopted search space, neural architectures are built stacking building cells upon the base cell. Every architecture starts with a pre-processing cell and ends with a classifier cell and can be conveniently represented by the tuple $(k, c)$, where $k$ is the number of kernels used in the base cell and $c$ is the number of cells stacked upon the base cell.

To explicitly take into account the limitations imposed by the device on which the HW-NAS should run, a custom constraint was added to the optimization problem in \cite{garavagno2024running} \cite{garavagno2024affordable}. Thus, a limit was added on the available memory for the training phase on the device hosting the search procedure. Accordingly, candidate architectures not fitting such constraint were considered unfeasible, even if they could be deployed on the target hardware.

\section{Proposal} \label{sec:proposal}
The core feature of the proposed HW-NAS is the capability to be ``hardware aware'' both at the edge (i.e., the device on which the NN should run) and at the gateway (i.e., the device on which the NAS should run). To this purpose, the HW-NAS involves six different constraints:   
\begin{itemize}
\item at the edge: a candidate solution (i.e., an architecture $A(k,c)$) must meet the constraints on the amount of RAM ($\xi_{RAM}$), the amount of Flash memory ($\xi_{Flash}$), and the MAC instructions ($\xi_{MAC}$). 
\item At the gateway: the whole process to be completed to identify the best architecture must meet the constraints on the amount of memory ($\xi_{MEM}$), the execution time ($\xi_{Time}$), and the energy budget ($\xi_{Energy}$).
\end{itemize}

In practice, one wants to keep the search space as wide as possible (i.e., the number of candidate solutions), given the constraints at the gateway level. The present approach generates the set $S_{\alpha}$ of admissible solutions with a two-step procedure. First, one selects the architectures $A(k,c)$ that 1) can be deployed on the edge device and 2) can be trained on the gateway without violating the constraint on the memory usage ($\phi_{RAM}(A) \leq \xi_{RAM}$): 

\begin{equation} \label{eq:problem2}
\begin{aligned}
\begin{cases}
   \hfil \forall A \in S_{\alpha}\\
   \hfil \phi_{RAM}(A) \leq \xi_{RAM} , \phi_{Flash}(A) \leq \xi_{Flash} \\
   \hfil \phi_{MAC}(A) \leq \xi_{MAC} , \phi_{MEM}(A) \leq \xi_{MEM} \\
   \hfil \xi_{RAM}, \xi_{Flash}, \xi_{MAC}, \xi_{MEM} > 0
\end{cases}
\end{aligned}
\end{equation}

Then, a cropping process reduces the cardinality of $S_{\alpha}$, if necessary. In fact, the search process must also meet the constraints on execution time and energy budget, which are evaluated by the functions $\phi_{Time}$ and $\phi_{Energy}$, based on the candidate search space. Hence, the number of candidate solutions in $S_{\alpha}$ should be as large as possible, but, at the same time, it should be small enough to allow the gateway to finalize the search process without violating the constraints $\xi_{Time}$ and $\xi_{Energy}$. Thus, the final set $S_{\alpha}$ is obtained as follows:

\begin{equation} \label{eq:problem1}
\begin{aligned}
\begin{cases}
   \hfil \max | S_{\alpha} |\\
   \hfil \phi_{Time}(S_{\alpha}) \leq \xi_{Time} , \phi_{Energy}(S_{\alpha}) \leq \xi_{Energy} \\
   \hfil \xi_{Time}, \xi_{Energy} > 0
\end{cases}
\end{aligned}
\end{equation}

Once the set $S_{\alpha}$ has been obtained, the search strategy can be applied to it. In the following, Sec. \ref{sec:adaptation_strategy} gives details about the algorithms that are adopted to generate $S_{\alpha}$, while Section \ref{sec:search_strategy} presents the search strategy.

\subsection{Generating the Search Space On The Gateway} \label{sec:adaptation_strategy}
The actual search space $S_{\alpha}$ is obtained after a two-step procedure. In the first step, Algorithm \ref{alg:adaptation1} builds the set $\hat{S}_{\alpha}$, i.e., a superset of $S_{\alpha}$, in that it should include all admissible architectures $A(k,c)$ given the deployment target (as per (\ref{eq:problem2})). It starts from the smallest admissible candidate, i.e., $(k = 1, c = 0)$, and explores sequentially all the candidate architectures $(1, c)$, adding them to $\hat{S}_{\alpha}$ until the first unfeasible architecture is reached (while loop at line \ref{alg:inner_level1} of the algorithm). Here, an architecture $A(k,c)$ is unfeasible when it does not meet the constraints set by (\ref{eq:problem2}). Then, $k$ is incremented (repeat loop at line \ref{alg:outer_level1} of the algorithm) and again the exploration along the $c$ parameter is completed. The algorithm continues accordingly until it reaches an architecture $A(k, 0)$ that is unfeasible (stopping criterion at line \ref{alg:stopping_criterion} of the algorithm).

\begin{algorithm}
\begin{algorithmic}[1]
\Require $\xi_{MEM}, \xi_{RAM}, \xi_{Flash}, \xi_{MAC}$  
\Ensure $\hat{S}_{\alpha}$
\Procedure{Search Space - Creation} {}
\State $k\gets 1$
\State $\hat{S}_{\alpha} \gets \emptyset$
\Repeat \label{alg:outer_level1} \Comment{Outer loop}
\State $c\gets 0$
\While{$A(k,c)$ is feasible} \label{alg:inner_level1} \Comment{Inner loop}
\State $\hat{S}_{\alpha}\gets \hat{S}_{\alpha} \cup (k,c)$
\State $c\gets c+1$
\EndWhile
\State $k\gets k+1$
\Until{$(k,0)$ is not feasible} \label{alg:stopping_criterion} \Comment{Stopping criterion}
\State \Return $\hat{S}_{\alpha}$
\EndProcedure
\end{algorithmic}
\caption{The algorithm that generates the extensive search space $\hat{S}_{\alpha}$}\label{alg:adaptation1}
\end{algorithm}

\begin{algorithm}
\begin{algorithmic}[1]
\Require $\hat{S}_{\alpha}$, $\xi_{Time}, \xi_{Energy}, \bar{w}$ 
\Ensure $S_{\alpha}$
\Procedure{Search Space - cropping}{}
\State $A^*\gets({A\in \hat{S}_{\alpha} | max\_params})$
\State $\bar{t}\gets eval\_time(A^*)$ \Comment{Time upper bound} \label{alg:timeupperbound}
\State $\bar{e}\gets \bar{t}\bar{w}$ \Comment{Energy upper bound} \label{alg:energyupperbound}
\State $k\gets 1$
\State $S_{\alpha}\gets \emptyset$ 
\State $stop \gets 0$
\Repeat \label{alg:outer_level2}
\State $c\gets 0$
\While{$(k,c) \in \hat{S}_{\alpha}$} \label{alg:inner_level2} 
\If{$|S_{\alpha}|\bar{t} \leq \xi_{Time} , |S_{\alpha}|\bar{e} \leq \xi_{Energy}$} \label{alg:feasibility1} 
\State $S_{\alpha}\gets S_{\alpha} \cup A(k,c)$
\Else 
\State {$stop \gets 1$}
\EndIf
\State $c\gets c+1$
\EndWhile
\State $k\gets k+1$
\Until $stop=0$
\State \Return $S_{\alpha}$
\EndProcedure
\end{algorithmic}
\caption{The algorithm that obtains the final search space $S_{\alpha}$ by cropping $\hat{S}_{\alpha}$.}\label{alg:adaptation2}
\end{algorithm}

Once $\hat{S}_{\alpha}$ has been computed, a cropping process can start. The issue to be tackled is that the gateway must meet the constraints on the execution time and the energy budget when running the NAS (as per (\ref{eq:problem1})). Roughly, both quantities depend on the number of architectures included in the search space, as the performance of each architecture should be assessed to find the best CNN among the candidate solutions. This in turn means that each architecture should be trained. Therefore, one cannot assume that the search space can be as large as $\hat{S}_{\alpha}$. At the start of the cropping process $S_{\alpha}$ coincides with $\hat{S}_{\alpha}$. Then, the cropping strategy formalized in the algorithm \ref{alg:adaptation2} basically reduces the cardinality of $S_{\alpha}$ to the purpose of obtaining a search space that allows the gateway to fit the constraints.  

First, an upper bound $\bar{t}$ is set for the time needed to evaluate a candidate architecture. To estimate such upper bound the algorithm selects the architecture with the highest number of parameters to be updated during training among those included in $\hat{S}_{\alpha}$. Accordingly, the value of $\bar{t}$ is set empirically by completing the evaluation process of such architecture (i.e., training and computation of the validation accuracy) on the gateway. The upper bound on the energy consumption, $\bar{e}$, is obtained by multiplying $\bar{t}$ by the maximum power consumption of the gateway when running the HW-NAS, $\bar{w}$. The latter quantity is an input to the algorithm. 

Then, starting from the smallest solution included in $\hat{S}_{\alpha}$, $(k = 1, c = 0)$, the algorithm adds one architecture at a time to $S_{\alpha}$ following the order in which the architectures have been added to $\hat{S}_{\alpha}$. The process ends as soon as the cardinality of $S_{\alpha}$ reaches  the threshold, i.e., when it is estimated that the gateway would not be able to complete the NAS by meeting the constraints on $\xi_{Time}$ and $\xi_{Energy}$. 

\begin{figure*}
    \centering
    \begin{subfigure}{0.3\textwidth}
        \centering
        \includegraphics[width=\textwidth]{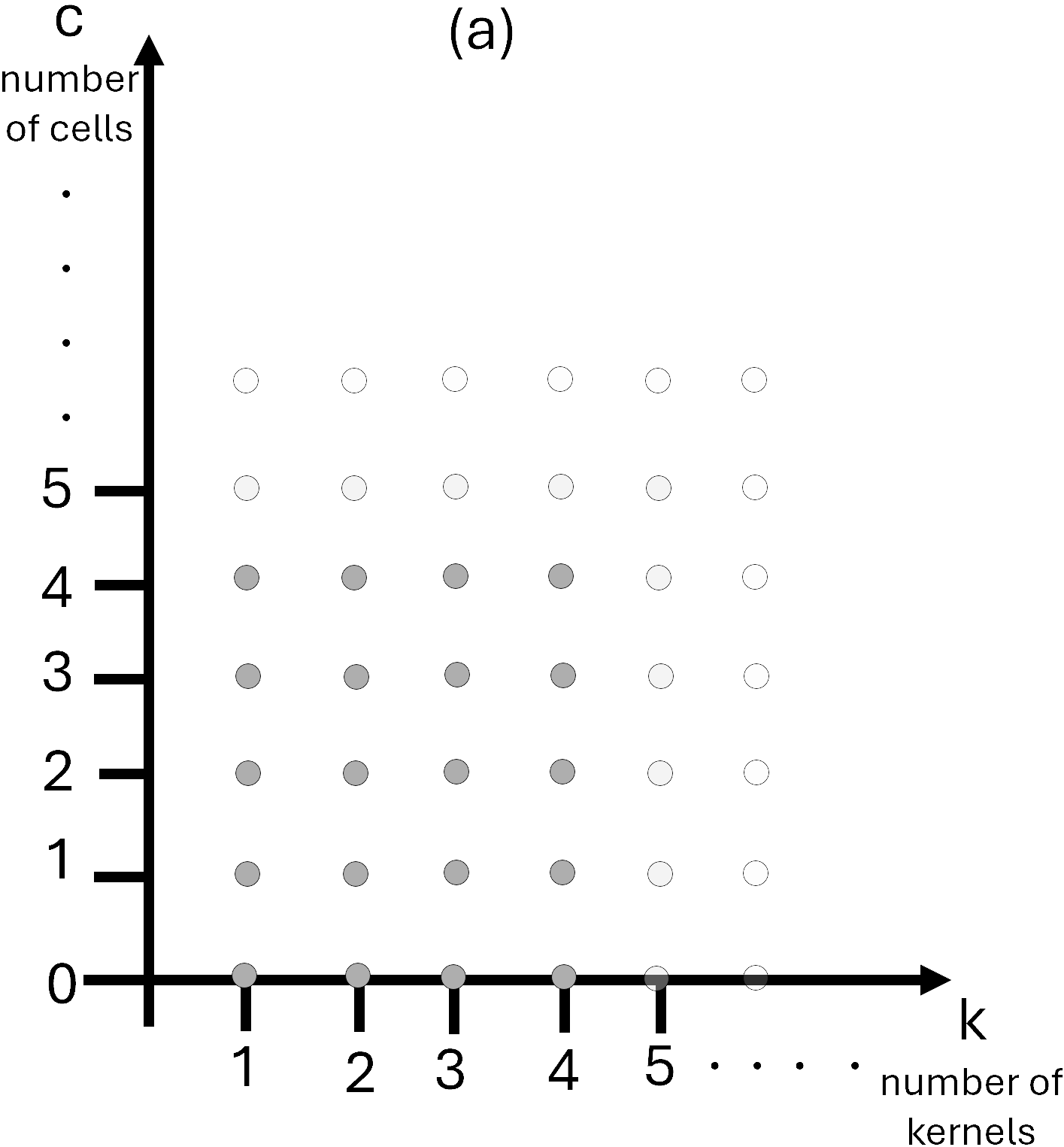}
        \subcaption{Hypothetical search space}
        \label{fig:searchspace_a}
    \end{subfigure}%
    ~ 
    \begin{subfigure}{0.3\textwidth}
        \centering
        \includegraphics[width=\textwidth]{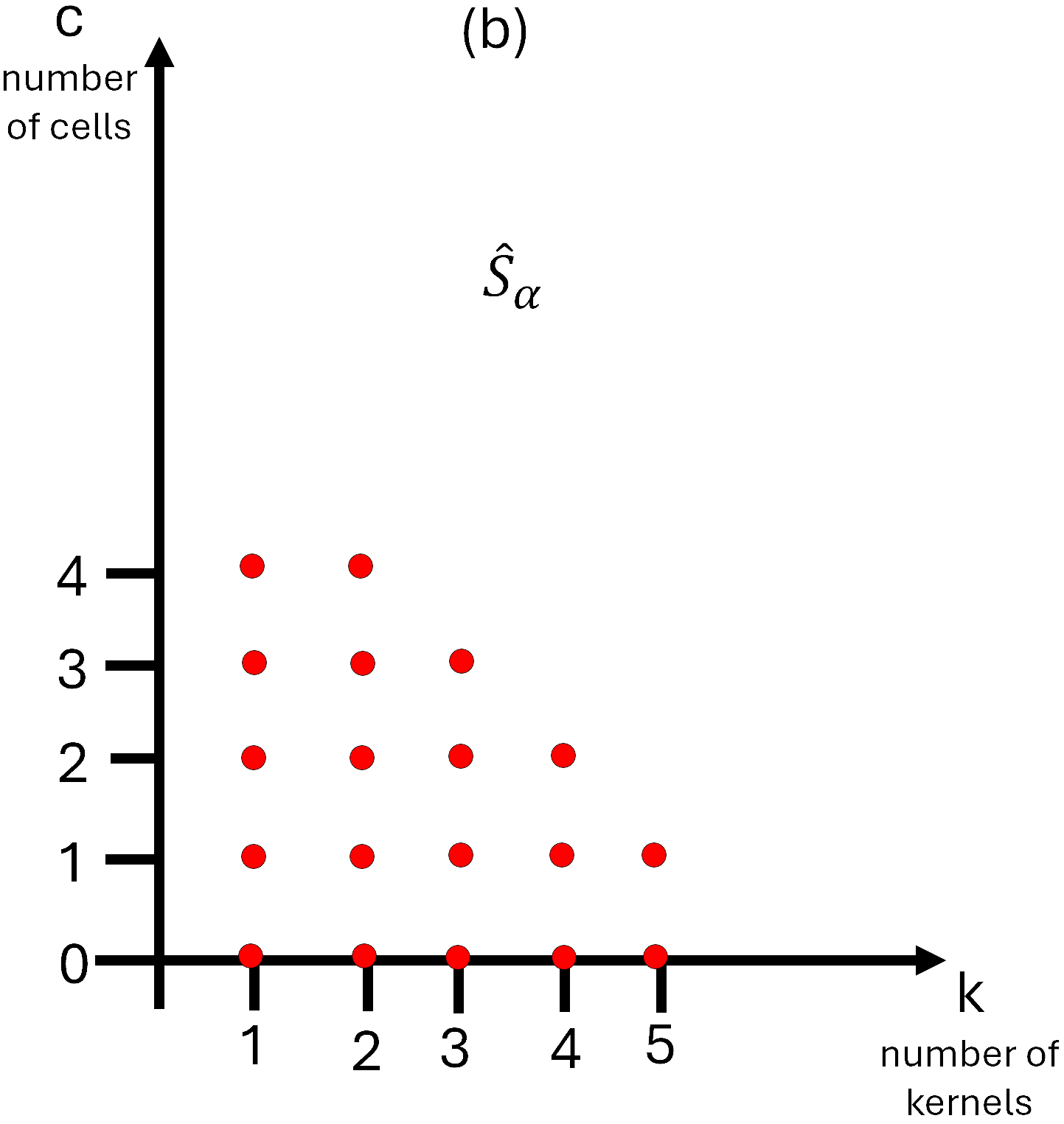}
        \subcaption{Extensive search space}%
        \label{fig:searchspace_b}
    \end{subfigure}
    ~ 
    \begin{subfigure}{0.3\textwidth}
        \centering
        \includegraphics[width=\textwidth]{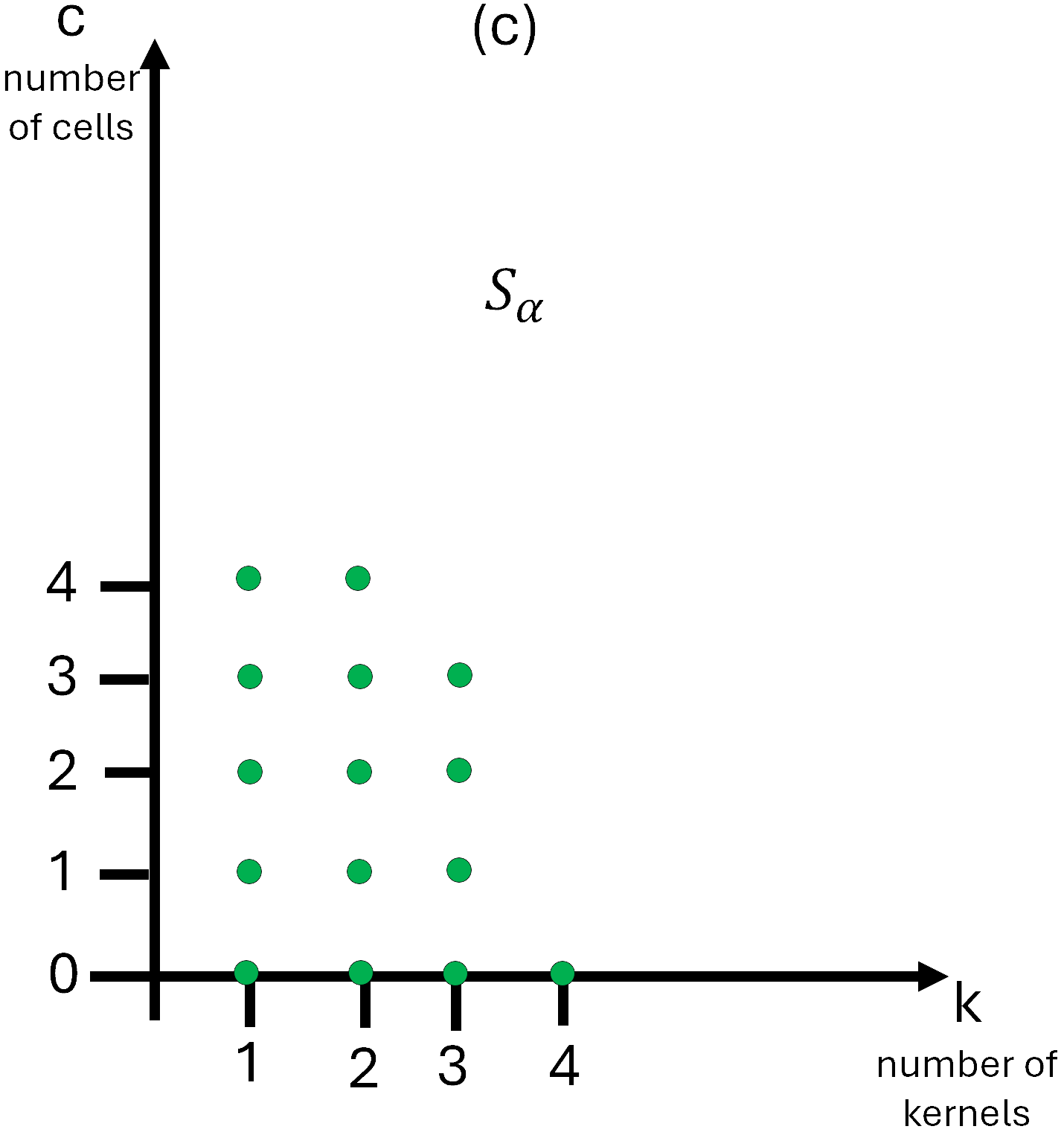}
        \subcaption{Actual search space}
        \label{fig:searchspace_c}
    \end{subfigure}
    \caption{Example of the process leading to the search space $S_{\alpha}$}
    \label{fig:searchspace}
\end{figure*}

Figure \ref{fig:searchspace} provides an example of the process leading to $S_{\alpha}$. The starting point is the hypothetical search space (Fig. \ref{fig:searchspace_a}), with all the possible architectures $A(k,c)$. The algorithm \ref{alg:adaptation1} generates an actual, yet temporary, search space $\hat{S}_{\alpha}$ (Fig. \ref{fig:searchspace_b}). Finally, the algorithm \ref{alg:adaptation2} yields as output $S_{\alpha}$ by cropping $\hat{S}_{\alpha}$ (Fig. \ref{fig:searchspace_c}).   

\subsection{Search Strategy} \label{sec:search_strategy}
The proposed search strategy is a bi-level optimization technique that does not use derivatives. It searches for the best neural architecture within the search space described in section \ref{sec:search_space} to solve the given classification task on the chosen microcontroller, with a preference towards smaller solutions. The inner level of the technique (Alg. \ref{alg}, while loop in row \ref{alg:inner_level}) searches the best number of building cells to stake given the number of kernels used in the base cell, acting on the parameter $c$, while the outer level (Fig. \ref{alg}, repeat until loop in row \ref{alg:outer_level}) changes the parameter $k$, to find the best combination of $k$ and $c$. 

\begin{algorithm}
\begin{algorithmic}[1]
\Procedure{proposed optimization technique}{}
\State $k\gets 1$, $c\gets 0$ \Comment{Starting point}
\State $\beta\gets 0$, $\gamma\gets 0$
\State $\bar{k}\gets k$
\Repeat \label{alg:outer_level} \Comment{Outer loop}
\State $\bar{c}\gets 0$, $c^{*}\gets 0$
\While{$(\bar{k},\bar{c}+1) \in S_{\alpha}$} \label{alg:inner_level} \Comment{Inner loop}
\If{$f(\bar{k},\bar{c}+1) > f(\bar{k},c^{*})$}
\State $c^{*}\gets \bar{c}+1$  \Comment{Update of best c}
\EndIf
\State $\bar{c}\gets \bar{c} + 1$
\EndWhile
\If{$f(\bar{k},c^{*}) > f(k,c)$} 
\State $k\gets \bar{k}$, $c\gets c^{*}$ \Comment{Candidate confirmation}
\Else
\State $\gamma\gets 1$ \Comment{Start k increment reduction}
\EndIf
\State $\beta\gets \beta + \gamma$ \Comment{Optional k increment reduction}
\State $\bar{k}\gets k + \lfloor 2^{-\beta}k \rfloor$ \Comment{Variable increment of k}
\Until{$\lfloor 2^{-\beta}k \rfloor = 0$} \Comment{Stopping criterion}
\EndProcedure
\end{algorithmic}
\caption{Proposed search strategy.}\label{alg}
\end{algorithm}

In detail, the inner level finds $c^{*}$, i.e., the best value of $c$ given a fixed $\bar{k}$ by adding building cells, one at a time, until the solution remains feasible. Instead, the outer level proposes a new candidate $\bar{k}$, based on the previously confirmed one $k$, obtained by adding to it a variable increment $\lfloor k2^{-\beta}\rfloor$, with beta initialized to zero, and calls the inner level. If the new candidate solution found is better than the previously confirmed one, the new candidate is confirmed, and the algorithm naturally proceeds. Otherwise, the beta starts to grow by one unit, decreasing the increment of the variable, and then a new $\bar{k}$ is proposed.  Eventually, the increment becomes zero, and the new proposed value coincides with the last confirmed one. When the latter condition is met, the algorithm stops, returning the last confirmed candidate.
The evaluation metric adopted is the generalization capability, i.e. the validation accuracy, computed over a validation set. 

\begin{figure}
    \centering
    \includegraphics[width=\linewidth]{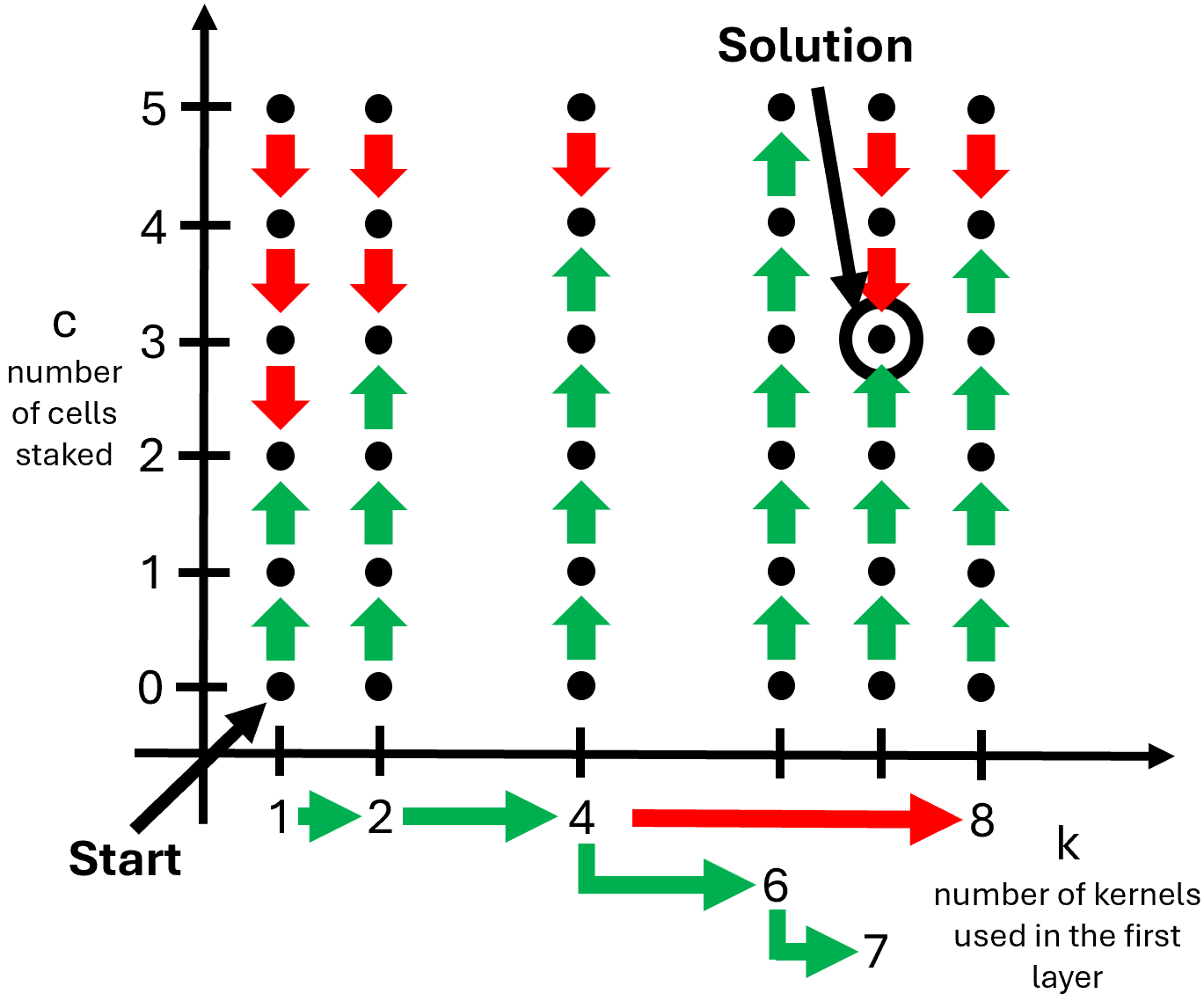}
    \caption{A possible run of the search strategy plotted on the search plane.}
    \label{fig:search_strategy_example}
\end{figure}

Figure \ref{fig:search_strategy_example} proposes a possible run of the search strategy plotted on the search plane. In this figure, every parameter change that results in a better candidate is marked with a green arrow, whereas changes that lead to a worse candidate are marked with red arrows.

The search starts from the smallest admissible solution, i.e. $A_{0} = A(k\gets 1, c\gets0)$ and explores the corresponding c-axis, finding the first candidate $A_{1} = A(\bar{k}\gets1, c^{*}\gets2)$. Since $A_{1}$ performs better than $A_{0}$, $A_{1}$ is confirmed, and the search proceeds by exploring the k-axis, doubling the confirmed k value, obtaining $\bar{k} = 2$. Then the c-axis is again explored finding the second candidate $A_{2} = A(\bar{k}\gets2, c^{*}\gets3)$. The new candidate is confirmed since it performs better than the previous one, i.e. $A_{1}$, and for this reason the change on the k-axis is marked with a green horizontal arrow. 

The algorithm continues in the same way, until the performance on the k-axis stops improving, in this case with $\bar{k} = 8$, which is indeed pointed by a red arrow. The correspondent candidate $A_{4} = A(\bar{k}\gets8, c^{*}\gets4)$ is therefore not confirmed leaving $A_{3} = A(k\gets4, c\gets4)$ in the position of the confirmed candidate. 

The stop of performance improvement triggers the k increment reduction, setting $\gamma\gets 1$. Therefore a new k value $\bar{k}\gets6$ is proposed. The corresponding candidate found by exploring the c-axis, i.e., $A_{5} = A(\bar{k}\gets6, c^{*}\gets5)$ performs better than $A_{3}$, the previous confirmed candidate, and therefore it is confirmed being marked with a horizontal green arrow pointing to it. 

The next candidate found $A_{6} = A(k\gets7, c\gets3)$ improves the latest confirmed solution and therefore is marked with a horizontal green arrow. The last k value proposed is 7, which is the same as the last confirmed solution because the variable increment reached the zero value. Therefore the algorithm terminates returning $A_{6}$.

\section{Experimental Setup} \label{sec:exp_setup}
This paper presents the outcomes of four different experiments:
\begin{itemize}
    \item The first experiment demonstrates that the proposed HW-NAS is capable of generating compact neural network architectures tailored for ultra-low-power microcontrollers with varying hardware capabilities.
    \item A second experiment compares the proposed HW-NAS with state-of-the-art methods, showing that it can deliver cutting-edge tiny CNNs optimized for ultra-low-power computing platforms.
    \item A third experiment demonstrates that the proposed HW-NAS can successfully adapt the cost of the search procedure to the constraints on execution time and energy budget set by an embedded device.
    \item A fourth experiment proves that the proposed HW-NAS can run on different embedded devices.
\end{itemize}

\paragraph{Gateways} The proposed HW-NAS has been executed on three different single-board computers that can serve as IoT gateways: the Raspberry Pi 4 (RPi 4), Raspberry Pi 3 (RPi 3), and Raspberry Pi Zero 2 (RPiZ 2). Table \ref{tab:embedded_boards} summarizes the key features of the boards: SoC information, available RAM, and measured peak power consumption. To evaluate the maximum power consumption of each device, the ``stress-ng'' utility from Debian was used. The following command line forced full utilization of all CPU cores and of the main memory of the device: ``stress-ng --cpu 0 --cpu-method all --vm 0 --vm-bytes 100\% --vm-method all --verify -t 1m -v''. A power meter, directly connected to the power supply, was used to measure the maximum power consumption of each device. The same power meter was also used to monitor the energy consumption and execution time of the devices during the HW-NAS process.

\begin{table}[!h]
    \centering
        \caption{Hardware features of the embedded devices targeted in this work.}
        \label{tab:embedded_boards} 
        \begin{tabular}{c c c c} 
                        \toprule
                        \bf{Embedded Device} & \bf{SoC}      & \bf{RAM}     & \bf{Power Consumption}  \tabularnewline 
                        \bf{Raspberry Pi}    & \bf{Broadcom} & \bf{[GiB]}   & \bf{[W]}     \tabularnewline \hline
                        4 Model B            & BCM2711       & 4            & 5.6            \tabularnewline 
                        3 Model B            & BCM2837       & 1            & 4.3            \tabularnewline 
                        Zero 2               & BCM2710A1     & 0.5          & 2.8            \tabularnewline \bottomrule
        \end{tabular} 
\end{table}

\paragraph{Sensor Nodes} Three different microcontrollers from the ultra-low-power production line of ST Microelectronics—namely, the L010RBT6, U083RCT6, and L412KBU3—have been selected as computing units for the sensor nodes. Table \ref{tab:nucleo_boards} shows the available RAM, Flash memory, and CoreMark score for each microcontroller.

%NUCLEO-L010RB, NUCLEO-U083RC, NUCLEO-L412KB
\begin{table}[!h]
    \centering
        \caption{Hardware features of the MCUs targeted in this work.}
        \label{tab:nucleo_boards} 
        \begin{tabular}{c c c c c} 
                        \toprule
                        \bf{MCU}   & \bf{RAM}   & \bf{Flash} & \bf{CoreMark}  \tabularnewline 
                        \bf{STM32} & \bf{[kiB]} & \bf{[kiB]} & \bf{score}     \tabularnewline \hline
                        \multicolumn{4}{c}{ultra-low-power line}              \tabularnewline
                        L010RBT6   & 20         & 128        & 75             \tabularnewline 
                        U083RCT6   & 32         & 256        & 134            \tabularnewline 
                        L412KBU3   & 40         & 128        & 273            \tabularnewline \bottomrule
        \end{tabular} 
\end{table}

\paragraph{NAS} 
The proposed HW-NAS indeed implemented a stand-alone process to monitor the execution time and to estimate the energy consumption in real-time. Such a backup process prevents the HW-NAS to run when either the bound on the execution time or the bound on the energy budget is reached. In fact, when dealing with an IoT gateway, non-real-time Operating Systems (OS), such as Debian or Windows, introduce unpredictable overloads for maintaining the system, which cannot be easily estimated. As a result, the search space $S_{\alpha}$ returned by Alg. \ref{alg:adaptation2} might result too large if the goal is to meet the constraints $\xi_{Time}, \xi_{Energy}$. Therefore, the software implementation of the HW-NAS is set to generate as output an architecture even when is not possible to fully explore the search space $S_{\alpha}$ as expected.  

\paragraph{Hyper-parameters} 
The maximum validation accuracy of a candidate architecture is evaluated after three epochs of training, with a validation split of 30\% of the training data, a batch size of $16$, and a learning rate of $1 \times 10^{-3}$. The Adam optimizer \cite{kingma2014adam} is adopted for this process. In the case of time series the architectures were evaluated after fifty epochs of training. 

\paragraph{Deployment}
Both Quantization Aware Training (QAT) and Post Training Quantization (PTQ) are utilized to obtain 8-bit quantized models. The memory features, such as Flash and RAM occupancy, have been measured using the STM utility called "stm32tflm", which is included in the X-CUBE-AI software package. This utility provides specific measurements for running the obtained architecture on STM32 microcontrollers using the TFLite Micro runtime. Additionally, the proposed method computes MAC operations and architectural details, denoted as $(k,c)$.

The architectures selected by the proposed strategy are finally trained for $100$ epochs, $500$ in the case of time series, with a validation split of $10\%$, a batch size of $128$, and a learning rate of $1e^{-2}$, using the Adam optimizer and implementing validation-based checkpointing. The training data is augmented by applying random horizontal flips and random rotations within the range $[-\frac{2}{5}\pi;\frac{2}{5}\pi]$. 

The reported latency is measured on the NUCLEO-L432KC using the 'validation on target' feature provided by X-CUBE-AI. X-CUBE-AI version 8.1.0, along with STM32CubeIDE version 1.17.0, was used to program the boards. 

\paragraph{Datasets} Four different datasets have been considered for the experiments:
\begin{itemize}
    \item the Visual Wake Words dataset \cite{chowdhery2019visual}, which is a tinyML benchmark \cite{banbury2021mlperf} containing 123,000 images that identify whether a person is present in the image;
    \item the CIFAR-10 dataset, which is a tinyML benchmark \cite{banbury2021mlperf} consisting of 60,000 32x32 color images organized into 10 classes, with 6,000 images per class;
    \item the Melanoma Skin Cancer dataset \cite{melanoma}, as a case for HIoT applications, which consists of 10,000 images used to identify whether a sample is malignant or not;
    \item the Case Western Reserve University (CWRU) dataset \cite{smith2015rolling}, as a case for IIoT application, which consists of accelerometer data coming from rolling bearings with different injected faults.
\end{itemize}

The test accuracy of the resulting architectures is evaluated on the public test set of the dataset used. 
In the case of the Visual Wake Words dataset, only one-tenth of the training split was used to run the HW-NAS, while the entire training split was utilized to train the resulting architecture.

\section{Results} \label{sec:experiments}
\subsection{Adaptability to ultra-low-power microcontrollers} \label{exp:hardware_awareness}
The first experiment aims to show that the proposed HW-NAS can generate custom architectures for different ultra-low-power computing platforms, which may play the role of sensor nodes in IoT applications. The assumption in this case is that the HW-NAS can run on a standard computing platform, i.e., without constraints on the available resources.

Table \ref{tab:hardware_awareness} presents the results, which are organized into three subtables, one for each dataset. Each subtable gives -for a target microcontroller- the configuration of the selected architecture, the hardware requirements of the deployed model (specifically RAM, Flash memory, and MAC operations), the test accuracy, and latency for a single prediction.

\begin{table}[!h]    \centering
    \caption{Performance of the proposed HW-NAS when targeting ultra-low-power computing platforms.}
    \label{tab:hardware_awareness}
    \begin{tabular}{c |c |c c c|c c}
        \toprule
        \bf{Target MCU} & \bf{Arch.} & \bf{RAM}   & \bf{Flash} & \bf{MAC}  & \bf{Test Acc.} & \bf{Lat.} \tabularnewline 
        \bf{STM32}      & \bf{(k,c)} & \bf{[kiB]} & \bf{[kiB]} & \bf{[MM]} & \bf{[\%]}      & \bf{[ms]} \tabularnewline \hline
        \multicolumn{7}{c}{Visual Wake Words ($vi=$ 50x50)} \tabularnewline
        L010RBT6        & (3,4)      & 19         & 10.8       & 0.4       & 71             & 42        \tabularnewline 
        U083RCT6        & (5,5)      & 24.5       & 22.7       & 0.9       & 75.2           & 63.2      \tabularnewline 
        L412KBU3        & (8,3)      & 31         & 18.8       & 2         & 78.3           & 79.1      \tabularnewline \hline 
        \multicolumn{7}{c}{Melanoma Skin Cancer ($vi=$ 50x50)} \tabularnewline
        L010RBT6        & (3,5)      & 18.5       & 8.1        & 0.4       & 84.2          & 43.2      \tabularnewline 
        U083RCT6        & (6,4)      & 26.5       & 20.4       & 1.3       & 88.5           & 67.3      \tabularnewline 
        L412KBU3        & (9,4)      & 34         & 35.7       & 2.6       & 90             & 129.8     \tabularnewline \hline
        \multicolumn{7}{c}{CIFAR10 ($vi=$ 32x32)} \tabularnewline
        L010RBT6        & (6,5)      & 14         & 28.7       & 0.5       & 63.3           & 31.6      \tabularnewline 
        U083RCT6        & (9,4)      & 17         & 36         & 1.1       & 67.1           & 60        \tabularnewline 
        L412KBU3        & (13,4)     & 21.5       & 65.2       & 2.1       & 70.9           & 97.5      \tabularnewline \bottomrule
    \end{tabular}
\end{table}

Experimental outcomes prove that the features of the selected architectures scale with the available resources on the target platforms. As resources increase, the HW-NAS selects an architecture that requires more RAM, Flash memory, and MACs. The test accuracy improves accordingly, just as latency increases. 

\subsection{State of the art comparison} \label{exp:state_of_the_art}
The proposed method is compared with four state-of-the-art NAS techniques: MCUNet \cite{lin2020mcunet}, Micronets \cite{banbury2021micronets}, ColabNAS \cite{garavagno2024colabnas}, and NanoNAS \cite{garavagno2024affordable}. The first two studies focused on high-performance MCUs, while the last two targeted ultra-low-power MCUs. The comparison is carried out using the Visual Wake Words dataset \cite{chowdhery2019visual}, a standard benchmark for tiny machine learning visual applications \cite{banbury2020benchmarking}. 

To ensure a fair comparison, Table \ref{tab:sota_comp} lists the smallest models selected by NAS targeting high-performance MCUs \cite{lin2020mcunet} \cite{banbury2021micronets} and the largest models selected by NAS targeting ultra-low-power MCUs \cite{garavagno2024colabnas} \cite{garavagno2024affordable}. For the method proposed in this work, the largest architecture among those reported in Section\ref{exp:hardware_awareness} has been selected. For each NAS, the table gives test accuracy, RAM, Flash memory, and MAC operations.

The results show that MCUNet produces the network with the highest test accuracy, albeit with the largest architecture in terms of RAM, Flash memory, and MAC operations. In contrast, the proposed method achieves the second-best test accuracy while maintaining the lowest resource usage in terms of Flash and the second-lowest in terms of RAM and MAC operations. ColabNAS ranks third in test accuracy, while featuring larger RAM and Flash. NanoNAS is fourth in test accuracy while featuring the lowest RAM occupancy and MAC operations. Micronets occupy the last position, delivering the lowest test accuracy but being only slightly smaller than MCUNet in RAM and Flash usage. 

\begin{table}[!ht]
    \centering
    \caption{Comparison with existing HW-NAS on VWW dataset.}
    \label{tab:sota_comp}
    \begin{tabular}{c c c c c c}
       \toprule
       \multirow{2}{*}{\bf{Work}}                 & \bf{Acc}  & \bf{RAM}   & \bf{Flash} & \bf{MAC} \\
                                                  & \bf{[\%]} & \bf{[kiB]} & \bf{[kiB]} & \bf{[MM]} \\
        \hline
        MCUNet  \cite{lin2020mcunet}              & 87.4      & 168.5      & 530.5      & 6 \\
        Micronets \cite{banbury2021micronets}     & 76.8      & 70.5       & 273.8      & 3.3 \\
        ColabNAS \cite{garavagno2024colabnas}     & 77.6      & 31.5       & 20.83      & 2 \\
        NanoNAS  \cite{garavagno2024affordable}   & 77        & 28.5       & 23.7       & 1.3 \\
        Proposal                                  & 78.3      & 31         & 18.8       & 2    \\
        \bottomrule
        & & & &
    \end{tabular}
    \
\end{table}

\subsection{Running HW-NAS on energy and time constrained devices} \label{exp:energy_awareness}
This experiment aims to demonstrate that the proposed HW-NAS exploits an adaptive strategy, which can fit the available resource budget. A Raspberry Pi Zero 2 has been selected as the gateway for this experiment, while STM32L412KBU3 has been used as sensor node. 

First, the extensive search space $\hat{S}_{\alpha}$ for the STM32L412KBU3 device has been generated according to Algorithm \ref{alg:adaptation1}. Then, the energy budget and the time budget required by Raspberry Pi Zero 2 to get $S_{\alpha}=\hat{S}_{\alpha}$ have been empirically estimated. Hence, in practice, it was assessed that with an energy budget of 16.5 Wh and a time budget 
of 9:51 (expressed hereafter as hh:mm) one can avoid any cropping of $\hat{S}_{\alpha}$ (according to Algorithm \ref{alg:adaptation2}). 

Table \ref{tab:energy_awareness} reports the results  of three experiments in which the proposed HW-NAS run on the selected gateway -Raspberry Pi Zero 2- with as many different energy / time budgets. The first row of the table shows the outcomes of the first experiment, in which the available budget was 16.5 Wh for energy and 9:51 for time. In this table, the second column gives the actual cost of the search procedure when running on the gateway; the third column gives the actual size of $S_{\alpha}$ after cropping, expressed as percentage with respect to $\hat{S}_{\alpha}$; the fourth column gives the percentage of $S_{\alpha}$ that has been covered by implementing the search strategy formalized in Algorithm \ref{alg}. The second part of the table provides details about the architecture generated by the NAS: RAM usage, Flash Memory usage, MAC operations, test accuracy on the VWW dataset, and latency measured on the STM32L412KBU3 device. The second row of the table reports the outcomes of the experiment in which the available budget was cut down to 11.0 Wh for energy and 6:34 for time, i.e., two-third of the best possible budget. Analogously, the third row of the table reports the results of the experiment in which the available budget further reduced to 5.50 Wh for energy and 3:17 for time, i.e., one-third of the best possible budget.

\begin{table}[!t]
    \centering
    \caption{Performance of the proposed HW-NAS when running on Raspberry Pi Zero 2 with different energy/time budgets, with STM32L412KBU3 as a target.}
    \label{tab:energy_awareness}
    \setlength\tabcolsep{2.5pt} 
    \begin{tabular}{c c c c | c c c c c}
        \toprule
        \multicolumn{4}{c|}{NAS Details}                  & \multicolumn{5}{c}{Resulting Architecture}                                 \tabularnewline
        
        \bf{Budget} & \bf{Cost} & \bf{$S_{\alpha}$} & \bf{Expl.} & \bf{RAM}   & \bf{Flash} & \bf{MAC}  & \bf{Acc.} & \bf{Lat.} \tabularnewline 
        \bf{[Wh][time]}    & \bf{[Wh][time]} & \bf{[\%]}      & \bf{[\%]}  & \bf{[kiB]} & \bf{[kiB]} & \bf{[MM]} & \bf{[\%]}       & \bf{[ms]} \tabularnewline \hline
        %(7,3) %search space trimmered: 100%
        16.5 - 9:51 & 16.5 - 9:51 &    100       & 51     & 28.5       & 16.1       & 1.6       & 77.8           & 87.4      \tabularnewline %(7,3) 
        %(4,4) %search space trimmered: 33%
        11.0 - 6:34 & 11.0 - 6:30  &    33       & 98     & 21.5       & 13.1       & 0.7       & 73.1           & 43.3  \tabularnewline %(4,4)
        %(1,4) %search space trimmered: 15%
        5.50 - 3:17   &  5.41 - 3:17 &   15     &  95     & 18.5       & 7.2        & 0.1       & 66             & 24.6  \tabularnewline %(1,4)
        \bottomrule
    \end{tabular}
\end{table}

The table shows that, as expected, only when the available budget was, respectively, 16.5 Wh for energy and 9:51 for time, the search procedure could work on the entire search space $\hat{S}_{\alpha}$ without any cropping: in fact, $S_{\alpha}$ was 100\% of $\hat{S}_{\alpha}$. Indeed, the search strategy managed to explore only 51\% of $S_{\alpha}$ to find the best architecture. In the other two experiments, the exploration of the search space $S_{\alpha}$ was halted due to resource depletion. In the second experiment, the search procedure was stopped after the threshold on the available energy was reached; four minutes were still available in terms of time budget. $S_{\alpha}$ was 33\% of $\hat{S}_{\alpha}$ and, notably, the search strategy was able to explore 98\% of $S_{\alpha}$ due to its small size. In the last experiment, conversely, the search procedure was stopped after the threshold on the available time was reached. $S_{\alpha}$ was 15\% of $\hat{S}_{\alpha}$ and the search strategy was able to explore 95\% of $S_{\alpha}$ due to its very small size.   

Obviously, the three experiments led to as many different architectures to be deployed on the target device. The footprint and the latency of the selected architecture reduce as the available budget for running the HW-NAS drops. In fact, generalization performance also deteriorates. Overall, the results obtained with this experimental session demonstrate that the proposed HW-NAS can adapt to the available budget by adjusting the portion of the search space to be explored.

\subsection{Effect of different platform on HW-NAS execution} \label{exp:platform_awareness}
This experiment aims to demonstrate that the proposed HW-NAS can adapt its search strategy to the characteristics of the specific gateway. To this end, three different devices that can be used as gateways have been involved in the experiment: Raspberry Pi Zero 2, Raspberry Pi 3 Model B and the Raspberry Pi 4 Model B. Again, STM32L412KBU3 has been used as sensor node.

Table \ref{tab:platform_awareness} reports on the result obtained by running the HW-NAS on the three selected gateways always with the same energy/time budget: 16.5 Wh for energy and 9:51 for time. As stated in the previous section, this budget allows one to avoid any cropping of the search space $\hat{S}_{\alpha}$ defined by the STM32L412KBU3 microcontroller when working on a Raspberry Pi Zero 2. 

The table gives, for each experiment: the gateway; the actual cost of the search procedure on that gateway, the actual size of $S_{\alpha}$ after cropping, expressed as percentage with respect to $\hat{S}_{\alpha}$; the percentage of $S_{\alpha}$ that has been covered by implementing the search strategy formalized in Algorithm \ref{alg}. In addition, the second part of the table provides details about the architecture generated by the NAS: RAM usage, Flash Memory usage, MAC operations, test accuracy on the VWW dataset, and latency measured on the STM32L412KBU3 device. 

The results show that when the Raspberry Pi 3 was adopted as a gateway, the search process stopped after 5 hours and 4 minutes due to energy depletion. In fact, Raspberry Pi 3 is generally less efficient than Raspberry Pi 0 in terms of energy consumption. Consequently, the proposed HW-NAS procedure had to cut $\hat{S}_{\alpha}$: eventually, $S_{\alpha}$ was 38\% of the extensive search space. Conversely, the HW-NAS was able to run on the the Raspberry Pi 4 without depleting the available budget. In this case, $S_{\alpha}$ was 100\% of the extensive search space; therefore, no cropping was needed. When running on the Raspberry Pi 4, the proposed HW-NAS managed to explore 57\% of the search space $\hat{S}_{\alpha}$ in 3 hours and 25 minutes, while 9 hours and 51 minutes were required on the Raspberry Pi Zero to explore 51\% of $\hat{S}_{\alpha}$. 

These results demonstrate that the proposed HW-NAS is effective in adapting to different execution platforms. The HW-NAS can set the size of $S_{\alpha}$ according to the available budget and the features of the gateway. In addition, the search strategy (Algorithm \ref{alg}) can adequately explore both small search spaces and large search spaces. 

\begin{table}[!h]
    \centering
    \caption{Performance of the proposed HW-NAS with STM32L412KBU3 as a target platform, maintaining a fixed resource budget while varying the execution platforms}
    \label{tab:platform_awareness}
    \setlength\tabcolsep{1.5pt} 
    \begin{tabular}{c c c c | c c c c c}
        \toprule
        \multicolumn{4}{c|}{NAS Details}                                                                 & \multicolumn{5}{c}{Resulting Architecture}                        \tabularnewline
        \multirow{2}{*}{\bf{RPi}} & \bf{Cost} & \bf{$S_{\alpha}$} & \bf{Expl.} & \bf{RAM}   & \bf{Flash} & \bf{MAC}  & \bf{Acc.} & \bf{Lat.} \tabularnewline 
                                  & \bf{[Wh][time]}      & \bf{[\%]}      & \bf{[\%]}  & \bf{[kiB]} & \bf{[kiB]} & \bf{[MM]} & \bf{[\%]}       & \bf{[ms]} \tabularnewline \hline
        %(7,3) %search space trimmered: 100%
        Z2                        & 16.5 - 9:51 & 100        & 51         & 28.5       & 16.1       & 1.6       & 77.8            & 87.4      \tabularnewline %(7,3)
        %(4,3) %search space trimmered: 36%
        3                         & 16.5 - 5:04 &  38        & 98         & 21         & 9.5        & 0.6       & 72.7            & 44.5  \tabularnewline %(4,3)
        %(9,5) %search space trimmered: 100%
        4                         & 14.2 - 3:25 & 100        & 57         & 35         & 52.1       & 2.6       & 75.9            & 125.7  \tabularnewline %(9,5)
        \bottomrule
    \end{tabular}
\end{table}
\vspace{-5pt}

\subsection{Time Series}\label{sec:time_series}
This experiment aims to show that the proposed technique can also be applied to time series, leading to state-of-the-art results on the CWRU dataset \cite{smith2015rolling}. A Raspberry Pi 4 has been adopted as gateway and a STM32-L010RBT6 as target sensor node. The algorithm has been run without energy and time constraints. The resulting network has been compared with the state-of-the-art work published by Chen et al. \cite{chen2020improved}.

\begin{table}[!ht]
    \centering
    \caption{Resulting Architecture VS Reference Architecture}
    \label{tab:result_time}
    \setlength\tabcolsep{1.5pt} 
    \begin{tabular}{c | c c | c c c c c c}
    \toprule
    \multirow{3}{*}{\bf{Work}}   & \multicolumn{2}{c|}{Search Cost} & \multicolumn{6}{c}{Resulting Architecture}                      \\ 
                                 & \bf{Energy} & \bf{Time}          & \bf{\multirow{2}{*}{(k, c)}} & \bf{RAM}   & \bf{Flash} & \bf{MAC}  & \bf{Test Acc.} & \bf{Latency} \\ 
                                 & \bf{[Wh]}  & \bf{[hh]:[mm]}     &                              & \bf{[kiB]} & \bf{[kiB]} & \bf{[MM]} & \bf{[\%]}      & \bf{[ms]}    \\ \hline
    Proposal                     & 6.4       & 1:52               & (6, 4)                       & 13.5       & 12.9       & 0.6       & 99.5           & 34        \\ 
    \cite{chen2020improved}      & n/a         & n/a                & n/a                          & 66.5       & 163.4      & 0.2       & 99.3           & 38.2      \\ \bottomrule
    \end{tabular}
\end{table}

Table \ref{tab:result_time} reports on the results of this experiment. The table compares the neural network generated by the proposed HW-NAS with that adopted in \cite{chen2020improved}. The table gives RAM usage, Flash Memory usage, MAC operations, test accuracy on the CWRU dataset, and latency measured on the STM32-L010RBT6 device.

The HW-NAS proved capable of producing a tiny CNN that needs 4.9 times less RAM and 12.7 times less Flash than the one adopted in \cite{chen2020improved}. Notably, such CNN scored higher test accuracy while decreasing latency. The resulting architecture can run on an STM32-L010RBT6, i.e., an ultra-low-power MCU, and is therefore suitable for IIoT sensor nodes. 

The search procedure was carried out on a Raspberry Pi 4 for 1 hour and 52 minutes; the energy consumption was 6.4 Wh. The average CPU usage per core was 67.9\%  and the average RAM usage was 593 MB (out of 4 GB available). This result makes the proposed method compatible with existing services running in deployed gateways of IoT networks.

\section{Conclusion} \label{sec:conclusion}
Using the proposed method, NNs can be automatically designed at the edge, without the data leaving the site of collection. This enables the usage of NNs even in privacy-sensitive applications: customized and hardware-friendly NNs can be designed for personalized healthcare in the HIoT framework, maintaining the data inside the hospital; or for IFD and quality control in production plants in the IIoT framework, preserving industrial secrets that could be disclosed by the data.

A thorough experimental campaign proved that the proposed HW-NAS a) can run on embedded devices that frequently serve as gateway nodes in IoT networks and b) achieves state-of-the-art results when ultra-low-power MCUs are selected as computing source for sensor nodes.  

To the best of the authors' knowledge the literature does not provide approaches to the design of HW-NAS that could allow the search process to run on an embedded device. The only exception is a recent work \cite{garavagno2024affordable} published by the authors of this paper; however, the HW-NAS presented in this work improves on \cite{garavagno2024affordable} in that it can dynamically adapt to the features of the gateway, thus enabling its use in IoT networks. The proposed approach may further simulate the development of lightweight HW-NAS that can run on resource-constrained devices, thus opening new fields of applications. Future works may focus, for example, on training-free evaluation techniques to further reduce the search cost.

\section{Acknowledgment}
\scriptsize
 Project funded under the National Recovery and Resilience Plan (NRRP), Mission 4 Component 2 Investment 1.1 - Call for tender No. 1409 published on Sept 14, 2022 by the Italian Ministry of University and Research (MUR) funded by the European Union – NextGenerationEU - Project Title "LEARN - muLtimodal Edge computing-bAsed weaRable exoskeletoNs for assistance in daily life" – CUP: D53D23016200001, J53D23014090001 - Grant Assignment Decree No. 1383 adopted on September 01, 2023 by the Italian Ministry of University and Research (MUR).

\vspace{5pt}

\noindent \includegraphics[width=0.5\textwidth]{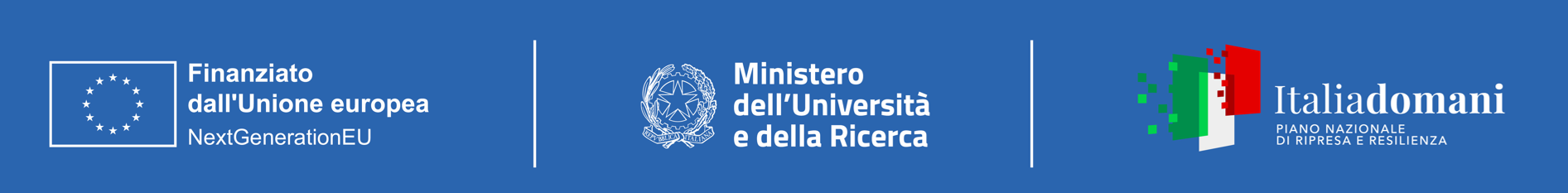}

\bibliographystyle{IEEEtran}
\bibliography{refs}

\end{document}